\documentclass[11pt]{article}
\usepackage{coling2020}
\usepackage{times}
\usepackage{url}
\usepackage{latexsym}

\colingfinalcopy

\usepackage{graphicx}
\usepackage{amsmath}
\usepackage{amssymb}
\usepackage{tikz}
\usepackage{multirow}
\usepackage{subcaption}
\usepackage{pifont}
\usepackage{makecell} 
\usepackage{wrapfig}
\newcommand{\cmark}{\ding{51}}%
\newcommand{\xmark}{\ding{55}}

\newcommand{\Fig}[1]{Fig.~\ref{#1}}

\newcommand{\Tbl}[1]{Tbl.~\ref{#1}}
\newcommand{\Sec}[1]{Sec.~\ref{#1}}

\newcommand{\benchmark}[1]{{\texttt{#1}}}
\renewcommand{\paragraph}[1]{\vspace*{0.05cm}\noindent\textbf{#1}\hspace*{.1cm}}

\newcommand*\circled[1]{\tikz[baseline=(char.base)]{
               \node[shape=circle,fill,inner sep=0.6pt] (char) {\textcolor{white}{#1}};}}

\title{How Far Does BERT Look At: \\Distance-based Clustering and Analysis of BERT's Attention}

\author{$\text{Yue Guan}^{1\star\dagger}$, $\text{Jingwen Leng}^{1\star\dagger}$, $\text{Chao Li}^{2\star\dagger}$, $\text{Quan Chen}^{2\star\dagger}$, $\text{Minyi Guo}^{2\star\dagger}$ \\
        $^\star$Shanghai Jiao Tong University\\
        $^\dagger$Shanghai Qi Zhi Institute\\
        ${ }^{1}\text{\texttt{\{bonboru,leng-jw\}@sjtu.edu.cn}}$\\
        ${ }^{2}\text{\texttt{\{lichao,chen-quan,guo-my\}@cs.sjtu.edu.cn}}$
}

\date{}

\begin{document}
\maketitle

\begin{abstract}
    Recent research on the multi-head attention mechanism, especially that in pre-trained models such as BERT, has shown us heuristics and clues in analyzing various aspects of the mechanism. As most of the research focus on probing tasks or hidden states, previous works have found some primitive patterns of attention head behavior by heuristic analytical methods, but a more systematic analysis specific on the attention patterns still remains primitive.
    In this work, we clearly cluster the attention heatmaps into significantly different patterns through unsupervised clustering on top of a set of proposed features, which corroborates with previous observations. We further study their corresponding functions through analytical study. In addition, our proposed features can be used to explain and calibrate different attention heads in Transformer models. 
\end{abstract}

\section{Introduction} \label{sec:introduction}

With the rapid development of neural network in NLP tasks these year, the Transformer~\cite{vaswani2017attention} that uses multi-head attention (MHA) mechanism is one recent huge leap \cite{goldberg2016primer}.
It has become a standard building block of recent NLP models.
The Transformer-based BERT~\cite{devlin2018bert} model further advances the model accuracy by introducing pre-training and has reached the state-of-the-art performance on many NLP tasks.

Beyond the general explanation which attributes the effectiveness of BERT model to its capability of long-range dependency and contextual embeddings, more detailed analysis of the MHA mechanism in BERT still remains an active research topic \cite{jain2019attention,serrano2019attention,ethayarajh2019contextual,michel2019sixteen}. 
\newcite{michael2020asking}; \newcite{brunner2020identifiability}; \newcite{coenen2019visualizing} explore the linguistic information or importance of BERT by inspecting hidden state embeddings. 
\newcite{wu2020lite}; \newcite{roy2020efficient} further investigate possible attention mechanism designs.

Several previous works on attention interpretability have analyzed the function and behavior of attention heads. 
\newcite{jawahar2019does} finds that attention heads within a same layer tend to function similarly, and vast redundancy could be found in the attention heads. \newcite{clark2019does}  provides the study on relating the behavior of each head to linguistic merits such as dependency parser. \newcite{kovaleva2019revealing} manually annotates attention heatmaps and trains a CNN to classify attention heads. Given these various analytical findings, from an algorithmic view, we are not aware of any research that provides a clear and reliable method to classify the attention heads without extra human intervention.

In this work, we provide a simple set of features that can be used to reliably disentangle the attention heads into different categories through clustering method. Different from heuristic analytics that looks at single inputs, our proposed attention head features are extracted corpus-wide and is thus more comprehensive in its discovered patterns. Interestingly, our algorithmically discovers patterns that match well with previous study~\cite{kovaleva2019revealing}. 
On the other hand, our method could easily be scaled up to large datasets and is generalizable to other similar multi-head-attention-based models.
With the use of the proposed clustering method, we further conduct empirical experiments to identify the important categories and parts of attention.
The results could explain previous MHA interpretability findings in terms of a distance view.

\section{Background and Motivation} \label{sec:background}

The BERT model~\cite{devlin2018bert} includes multiple layers of Transformer~\cite{vaswani2017attention} encoders.
Its core component is multi-head attention mechanism (MHA).
Given a sequence of input embeddings, the output contextual embedding is composed by the input sequence with different attention at each position.
The attention weight is calculated as following,
\begin{equation} \label{equ:attention}
    \small
    Attention_{(Q,K)} = Softmax(\frac{QK^T}{\sqrt{d_k}}),
\end{equation}
where $Q, K$ are query and key matrix of input embeddings, $d_k$ is the length of a query or key vector.
Multiple parallel groups of such attention weights, also referred as attention heads, make it possible to attend to information at different positions.
BERT uses special tokens to encode the different input and output formats uniformly for downstream tasks, such as \benchmark{\small [CLS]} and \benchmark{\small [SEP]}.

\begin{figure*}[t]
  \begin{subfigure}{.25\columnwidth}
    \centering
    \includegraphics[width=0.45\linewidth]{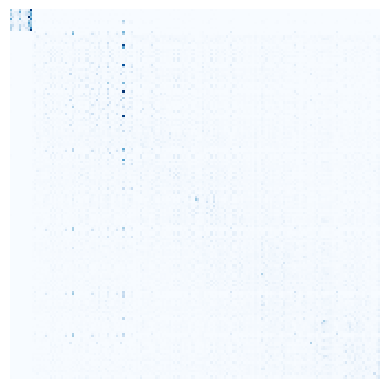}~
    \includegraphics[width=0.45\linewidth]{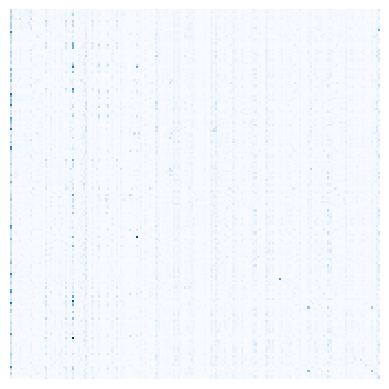}
    \includegraphics[width=\linewidth]{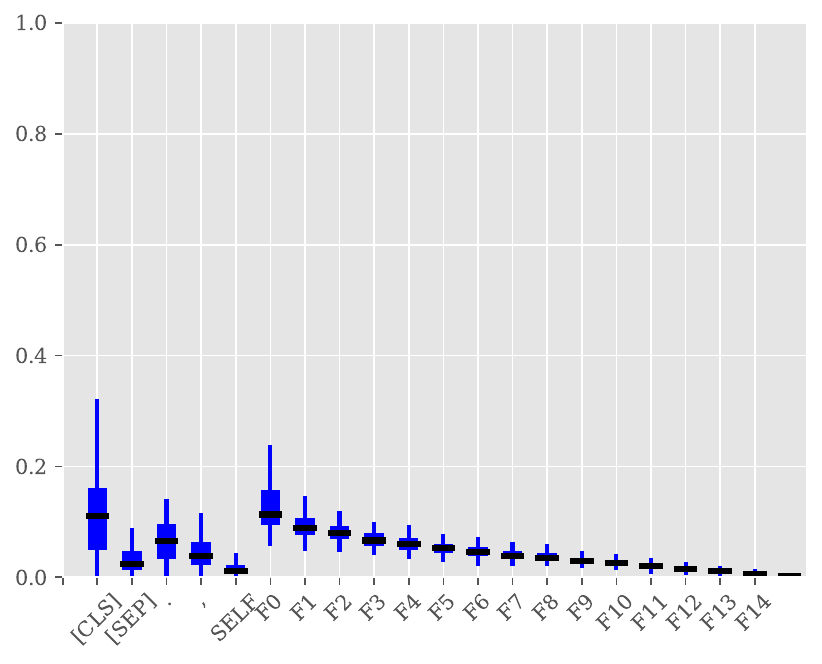}
    \caption{Dense type.}
    \label{fig:sub-first}
  \end{subfigure}
  \begin{subfigure}{.25\columnwidth}
    \centering
    \includegraphics[width=0.45\linewidth]{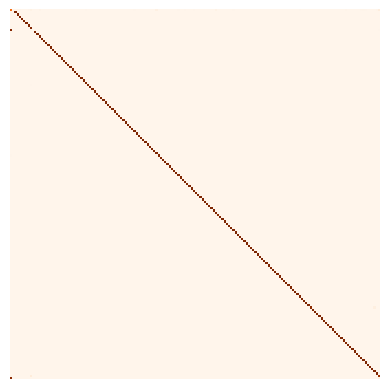}~
    \includegraphics[width=0.45\linewidth]{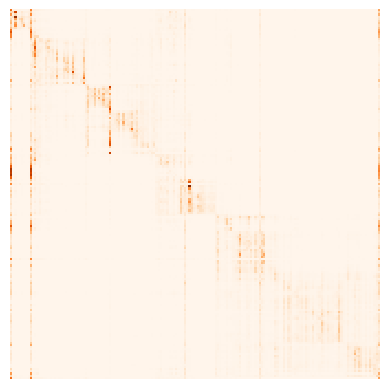}  
    \includegraphics[width=\linewidth]{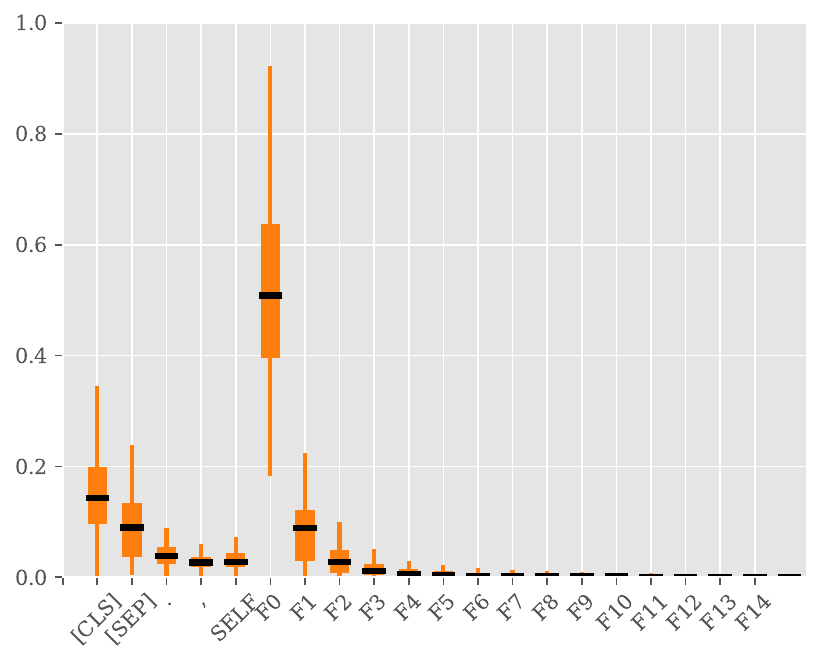}
    \caption{Diagonal type.}
    \label{fig:sub-first}
  \end{subfigure}~
  \begin{subfigure}{.25\columnwidth}
    \centering
    \includegraphics[width=0.45\linewidth]{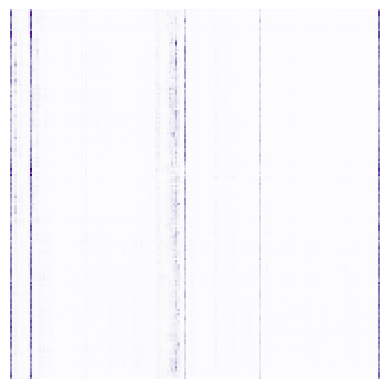}~
    \includegraphics[width=0.45\linewidth]{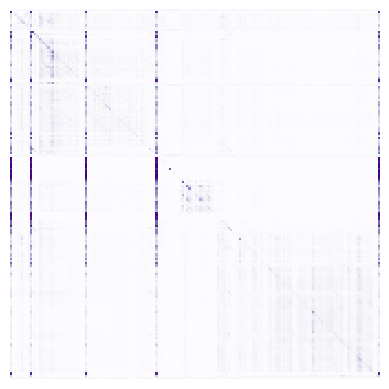}  
    \includegraphics[width=\linewidth]{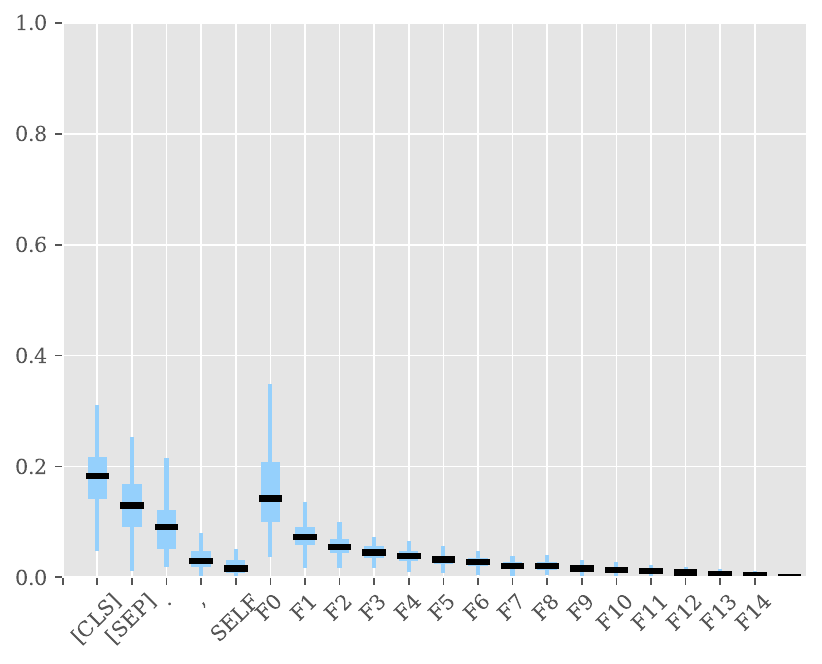}
    \caption{Dense\&vertical type.}
    \label{fig:sub-first}
  \end{subfigure}~
  \begin{subfigure}{.25\columnwidth}
    \centering
    \includegraphics[width=0.45\linewidth]{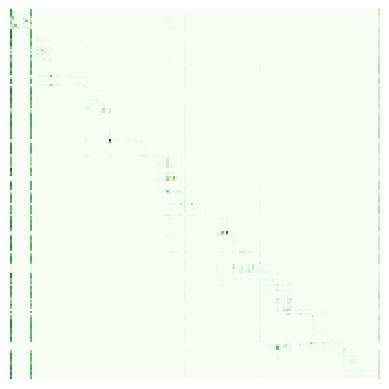}~
    \includegraphics[width=0.45\linewidth]{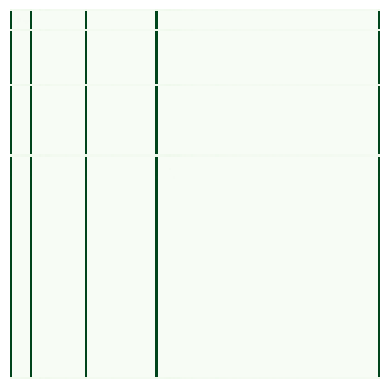}  
    \includegraphics[width=\linewidth]{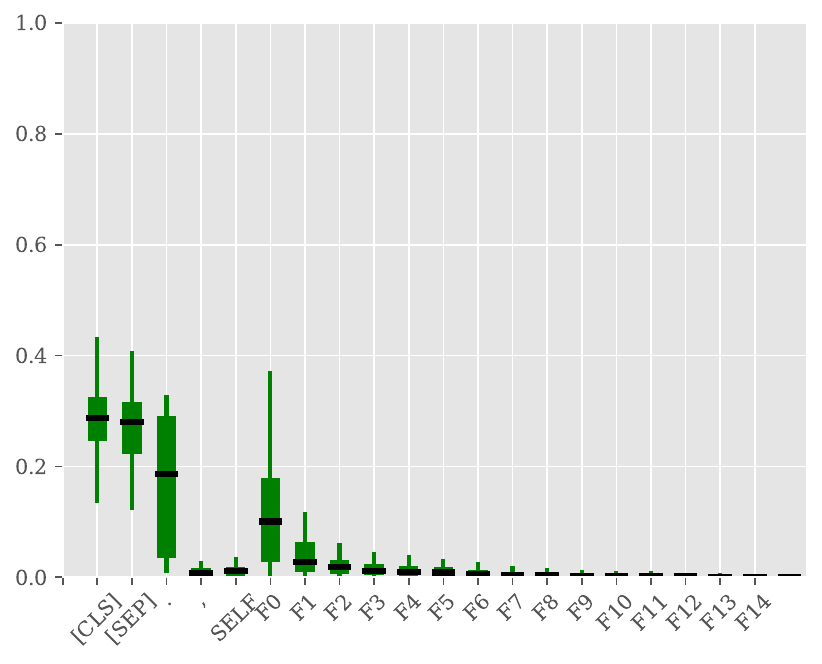}
    \caption{Vertical type.}
    \label{fig:sub-first}
  \end{subfigure}
  \caption{\small 
  We extract the distance features and perform the K-means clustering of 384 attention heads in the BERT-large model. Top: two examples in each attention type. Bottom: the box-plot of 21-dimensional distance features in each type.}
  \label{fig:heatmaps}
  \vspace*{-0.3cm}
\end{figure*}

Given the great success of BERT-like models, researchers start to explore the working mechanisms of multi-head attention component.
One common approach is to visualize the attention weight matrix~\cite{voita2019analyzing,clark2019does,kovaleva2019revealing}, which often finds that attention heads in BERT demonstrate several specific patterns.
For example, some attention heads show a dominant stripe pattern on diagonal direction while some others show a dominant vertical stripe pattern.
There are also attention heads with a relatively homogeneous distribution of attention weights.
In this work, we refer this patterns as dense, vertical or diagonal with some samples in the top part of \Fig{fig:heatmaps}.

However, these previous works rely on simple heuristic rules or human annotation, which makes them not scalable and error-prone. 
In this work, we propose a scalable and efficient unsupervised learning approach that automatically clusters different attention heads, which lets us systematically study their different roles in NLP tasks.

\section{Distance Feature} \label{sec:feature}

Based on the attention pattern findings in previous works, we design a distance feature that enables unsupervised clustering of attention heads.
For each input token, we first accumulate its attention weights to the special tokens including \texttt{\small '[CLS]'}, \texttt{\small'[SEP]'}, \texttt{\small','} and \texttt{\small '.'}, because the vertical stripe pattern shows strong focus on these tokens.
We then accumulate its attention weights to nearby tokens within a window size, which lets us distinguish the diagonal stripe with strong locality and dense pattern with weak locality.
For a input sample with length of $L$, we divide the sample to $N$ windows, each of which has a size of $L/N$.
Eq.~\ref{eq:feature} calculates the window feature $F(i)$ of $i_{th}$ window.
\begin{equation} \label{eq:feature}
    \small F(i)=\frac{1}{L}\sum_{s=1}^{L} \sum_{t=iL/N}^{(i+1)L/N}  Attention(s,s\pm t)
\end{equation}
The special token attention and window attention are exclusive and we normalize both them by the input sequence length $L$.
As such, their summation for a give input token remains one.

For an input sample, attention weight matrix from all tokens are averaged to get a low-dimensional representation of an individual head.
With 16 windows, we can represent an attention head with a 21 (16 windows + self + 4 special tokens) dimensional feature for further analysis.

\begin{figure}[t]
  
  \begin{minipage}{0.28\columnwidth}
    \centering
    \includegraphics[width=\columnwidth]{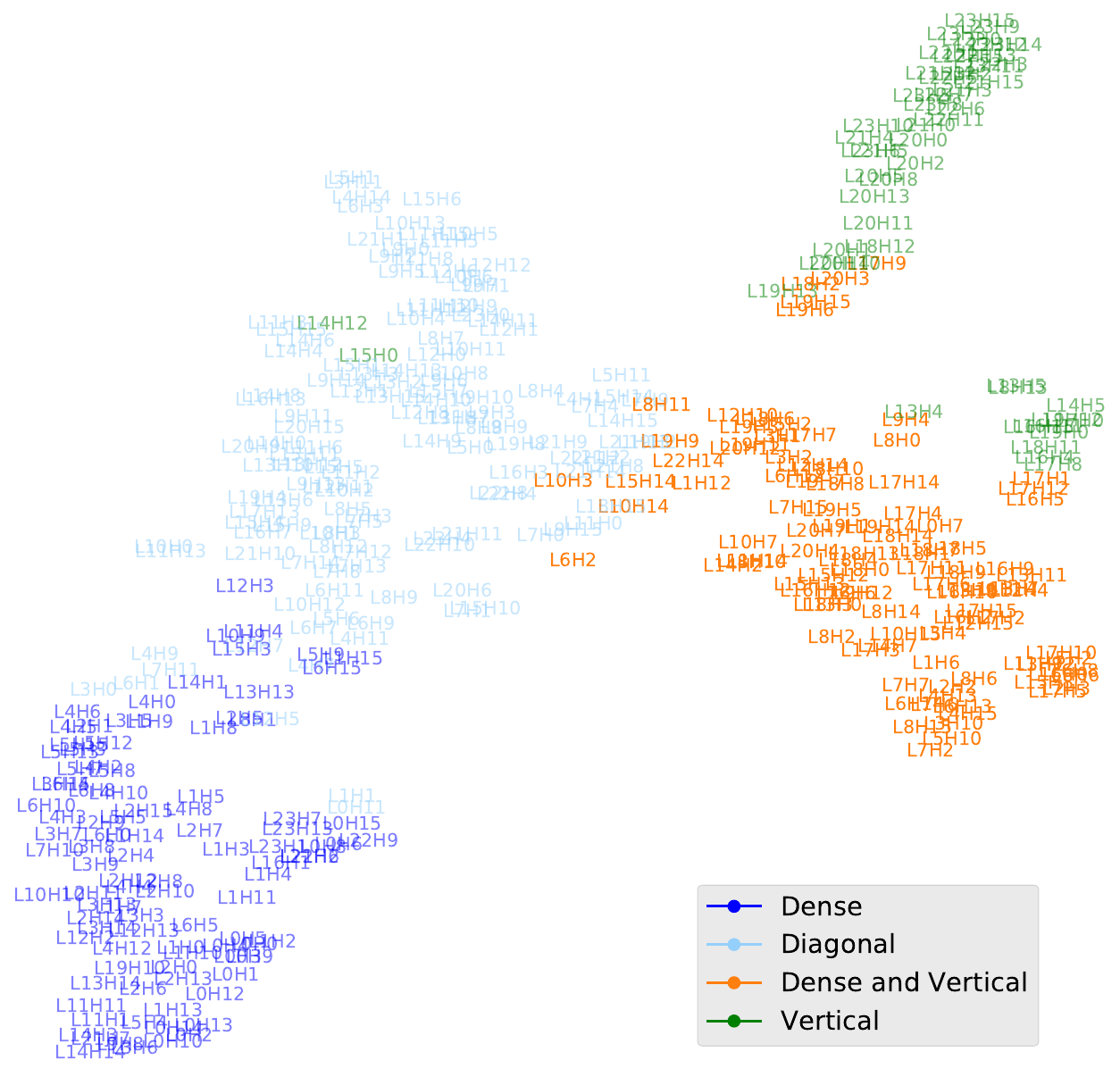}
    \caption{\small T-SNE visualization of K-means clustering with the distance feature.}
    \label{fig:tsne2}
  \end{minipage}~
  \begin{minipage}{0.34\columnwidth}
    \centering
    \includegraphics[width=\columnwidth]{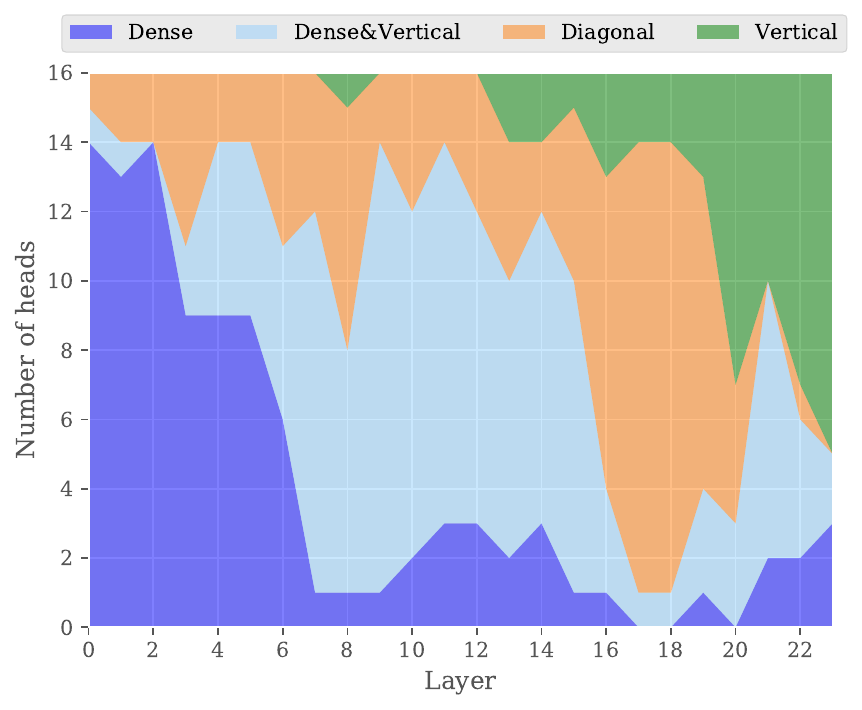}
    \vspace{-0.6cm}
    \caption{\small The attention head type distribution across different layers in the BERT-large model.}
    \label{fig:head_distribution}
  \end{minipage}~
  \begin{minipage}{0.38\columnwidth}
    \begin{subfigure}{.5\columnwidth}
      \centering
      \includegraphics[width=\linewidth]{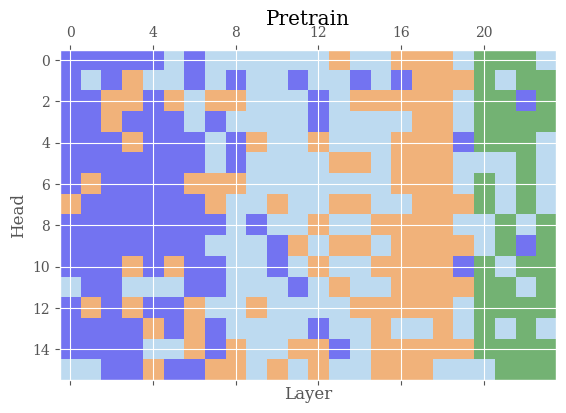}  
      \label{fig:sub-first}
      \vspace{-0.5cm}
    \end{subfigure}~
    \begin{subfigure}{.5\columnwidth}
      \centering
      \includegraphics[width=\linewidth]{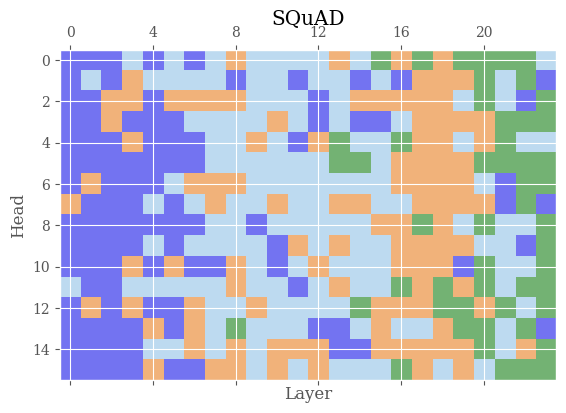}  
      \vspace{-0.5cm}
      \label{fig:sub-second}
    \end{subfigure}
    \begin{subfigure}{.5\columnwidth}
      \centering
      \includegraphics[width=\linewidth]{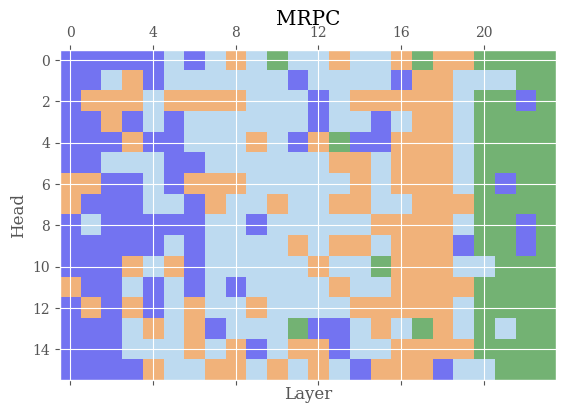}  
      \label{fig:sub-first}
      \vspace{-0.6cm}
    \end{subfigure}~
    \begin{subfigure}{.5\columnwidth}
      \centering
      \includegraphics[width=\linewidth]{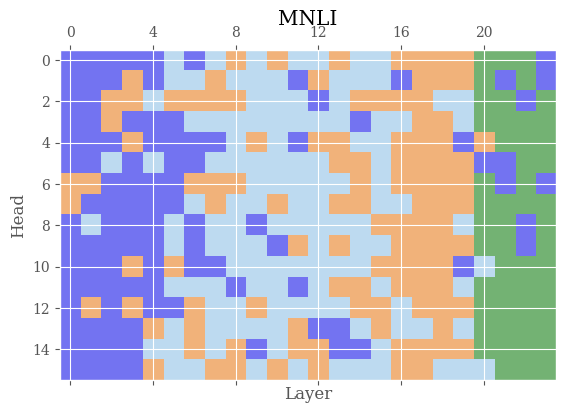}  
      \vspace{-0.6cm}
      \label{fig:sub-second}
    \end{subfigure}

    \caption{\small Clustering result on pre-trained language model and models fine-tuned on downstream tasks.}
    \label{fig:downstream}
  \end{minipage}
  \end{figure}

\section{Clustering} \label{sec:clustering}

We now describe how we leverage the 21-dimensional distance feature to apply an unsupervised attention head clustering.
We study attention behavior in the BERT-large model\footnote{\scriptsize We use \texttt{bert-large-uncased-whole-word-masking-finetuned-squad}~\cite{Wolf2019HuggingFacesTS} from https://huggingface.co/models.}.
We randomly pick 1,000 samples from the SQuAD dataset~\cite{rajpurkar2016squad} and extract the averaged distance feature across all selected samples, which is used to represent an attention head.

To verify if there exists clustering behavior in BERT, we first use a non-linear dimension reduction algorithm T-SNE~\cite{maaten2008visualizing} to visualize all 384 heads (24 layers $\times$ 16 heads per layer). 
The result in \Fig{fig:tsne2} clearly demonstrates four clusters.
We then use K-means clustering algorithm~\cite{hartigan1979algorithm} with 4 clusters and the result is shown as the colors.

\Fig{fig:heatmaps} shows two examples (top) and the distance feature box-plot (bottom) for each cluster. 
We also show the distribution of each type of attention head across different layers in \Fig{fig:head_distribution}.
Our results show that an attention head shows a consistent weight pattern across different input samples and there exists a strong clustering behavior across different layers.
We describe the key findings in details.

\textbf{\benchmark{Dense}:} This type has dense attention patterns (\Fig{fig:heatmaps}(a) top).
\Fig{fig:heatmaps}(a) bottom shows that the heads in this type have the highest attention at long range windows and relatively small attention weights to the special tokens.
They account for $26.04\%$ of total heads and mainly appear at the beginning layers.

\textbf{\benchmark{Diagonal}:} This type has dominant diagonal patterns and a few vertical stripes (\Fig{fig:heatmaps}(b) top).
Its distance features show large values in the short-range windows and medium values in the special tokens (\Fig{fig:heatmaps}(b) bottom).
The short-range attentions can capture the local information while the vertical stripe can carry the global information.
They account for $25.56\%$ of total heads and appear at almost all layers.

\textbf{\benchmark{Dense\&Vertical}:} This type mixes the dense and vertical patterns (\Fig{fig:heatmaps}(c) top). 
Its distance features have large values at long range windows and special tokens.
They account for $34.11\%$ of total heads and mainly appear at middle layers.

\textbf{\benchmark{Vertical}: }  This type has mostly strong vertical patterns (\Fig{fig:heatmaps}(d) top). 
Its distance features have large values at special tokens, which are much higher than other types.
They account for $13.28\%$ of total heads and mainly appear at higher layers.

As a matter of fact, it is possible to exploit the proposed distance feature with other unsupervised clustering algorithms.
Our work chooses to use the K-means clustering algorithm as a representative one.
With Gaussian mixture model clustering~\cite{amendola2015moment}, we also get similar results.
As such, we choose to use four clusters in the following experiment based on the T-SNE visualization result.
Moreover, we find the four clusters are human interpretable. 
We still get stable clustering results with 3 or 5 clusters, but the patterns can not be distinguished by human-being.

We measure the stability of clustering and generality on a pre-trained language model and downstream tasks including question answering, sentence pair similarity regression and natural language inference.
For the aforementioned tasks, we use datasets that include SQuAD~\cite{burger2001issues}, MNLI~\cite{N18-1101}, and MRPC~\cite{dolan2005automatically} respectively.
We also extract the features on fine-tuned BERT-large models accordingly.
All of these training settings are inline with \newcite{Wolf2019HuggingFacesTS} and \newcite{devlin2018bert}.

\begin{table}
  \centering
  \begin{tabular}{ccccc}
    \Xhline{1.5pt}
  &Pre-train & SQuAD   & MRPC    & MNLI    \\\hline
  Stability&95.98\%   & 96.26\% & 95.98\% & 94.37\% \\\Xhline{1.5pt}
  \end{tabular}
  \caption{Stability of clustering measured following \newcite{von2010clustering} on downstream tasks.}
  \label{tbl:stability}
\end{table}

We evaluate the stability of clustering following \newcite{von2010clustering}.
The idea is to obtain multiple models with partial training data and evaluate the consistent samples among all fitted models.
We randomly split $50\%$ training data to fit clustering models and predict on the whole dataset.
The stability of clustering algorithm is calculated as the proportion of input samples that is consistent on all models fitted with the random partial training set split.
We report the clustering stability of 10 times clustering in \Tbl{tbl:stability}.
Our result demonstrates that over $95\%$ attention heads are clustered stably with a limited amount of data. 
This shows that the proposed method is generally applicable and scalable.

The detailed results are shown in \Fig{fig:downstream}, which demonstrates similar type distribution among different models.
In specific, all the four models share similar head type distribution described in \Fig{fig:head_distribution}.
Empirically, we find many heads have the same cluster after fine-tuning process.
However, some heads in the highest layers also change.
We suggest that higher layers are more relevant to task specific knowledge while lower layers handles more general linguistic features.
Particularly, pre-trained model and SQuAD-fine-tuned model have more \texttt{dense\&vertical} types at highest layer, which we will show to be responsible for global information in \Sec{sec:experiments}.
This is explained by the fact that question answering and language modeling are more difficult tasks as they use token embeddings from all positions for classification.
On the contrast, the other two tasks only use the special token \benchmark{\small [CLS]}.

\section{Experiments} \label{sec:experiments}

Based on the attention clustering results, we now study which attention head type(s) is important and what range of information different head types process.
In specific, we manipulate the different attention heads in BERT-large model and observe the resulted accuracy impact on SQuAD v1.1 development set.
It should be mentioned that all these experiments are directly ablation without fine-tuning on the model parameters.

\subsection{Importance of Head Types} \label{sec:experiments:type}

We first prune or substitute the attention weight of a particular type to study its importance.
We mask the attention value to $0$ or substitute it with an uniformed attention value of $1/L$.
\Tbl{tbl:pruning} reports the experimental results with different settings.

Firstly, we find that pruning all \benchmark{diagonal} heads (\circled{3}) makes the model totally inoperative, compared with (\circled{2},\circled{4},\circled{5}).
Pruning all \benchmark{vertical} heads directly (\circled{5}) has less than $1.5\%$ accuracy loss.
This indicates that \benchmark{Diagonal} type is vital for BERT processing and \benchmark{vertical} type is the opposite.
On top of this, we further prune two types together with \benchmark{diagonal} type always remained.
Keeping \benchmark{dense\&vertical} type (\circled{6}) together results in a better accuracy.
Nextly, we find that substituting the \benchmark{dense} type with uniformed attention (\circled{8}) results in accuracy loss of less than $10\%$.
This indicates that the attention values of \benchmark{dense} are trivial.
Substituting \benchmark{Vertical} type ({\small \circled{11}}) also has an accuracy of over $50\%$.
However, substituting the mixed type \benchmark{dense\&vertical} ({\small \circled{10}}) has a significant accuracy loss, which indicates its crucial role.

\subsection{Information in Important Heads} \label{sec:experiments:part}

After identifying the \benchmark{diagonal} and \benchmark{dense\&vertical} types as the important heads, we continue to study which part of information they are responsible for in terms of attending distance.
We constrain attention matrix by sentences and remove the intra-sentence attention or inter-sentence attention.
When the intra-sentence attention is removed, the attention head can only access tokens in other sentences and obtains global information.
On the contrast, the attention head has only local syntactic or word level semantic information.
We remove intra or inter sentence attention of one type to study the impact on the accuracy ({\small \circled{12}}-{\small \circled{15}}).
Nextly, we keep intra or inter sentence attention of one type to see how much information is carried out by it ({\small \circled{16}}-{\small \circled{19}}).

\Tbl{tbl:pruning_distance} demonstrates the experimental results.
Removing intra-sentence attention from \benchmark{diagonal} type ({\small \circled{12}}) leads to a poor accuracy of $19.96\%$, while removing inter-sentence part ({\small\circled{13}}) has little interference.
This proves that \benchmark{diagonal} type focus on short distance information and such information is substantial for its important role in the model.
Even with only \benchmark{diagonal} head handling intra-sentence information ({\small \circled{16}}), the model has a descent accuracy of $71.80\%$.
On the contrast, with \benchmark{dense\&vertical} attending other sentences solely ({\small \circled{19}}), the model reaches an accuracy of $75.54\%$.
This result shows that the BERT model accesses to long distance information with \benchmark{dense\&vertical} mostly. 
This indicates \benchmark{diagonal} and \benchmark{dense\&vertical} type  are responsible for local and global information respectively.

Finally, we draw a general view about the attention behaviors according to our analytic results in \Sec{sec:experiments:type} and \Sec{sec:experiments:part}.
\benchmark{Diagonal} type is the most important and they focused purely on intra-sentence processing.
\benchmark{Dense\&Vetical} type absorb information globally from other sentences.
The \benchmark{vertical} type is insignificant.
Attention values of \benchmark{dense} type are trivial and is substituted by uniformed attention without large interferences. 

\begin{table}[]
  \begin{minipage}{.48\linewidth}
  \resizebox{\linewidth}{!}{
  \begin{tabular}{c|c>{\centering\arraybackslash}p{1.8cm}>{\centering\arraybackslash}p{1.8cm}>{\centering\arraybackslash}p{1.8cm}>{\centering\arraybackslash}p{1.8cm}|cc}
    \Xhline{1.5pt}
                            & id  & Dense         & Diagonal      & Dense\&Vertial & Vertical      & Acc.  & F1    \\ \hline
  Baseline                  &\circled{1}  & \cmark   & \cmark   & \cmark    & \cmark   & 82.85 & 89.68 \\ \Xhline{1.5pt}
  \multirow{6}{*}{Prune}    &\circled{2}  & \xmark & \cmark   & \cmark    & \cmark   & 79.92 & 87.62 \\ 
                            &\circled{3}  & \cmark   & \xmark & \cmark    & \cmark   & 4.59  & 7.55  \\ 
                            &\circled{4}  & \cmark   & \cmark   & \xmark  & \cmark   & 73.33 & 81.92 \\ 
                            &\circled{5}  & \cmark   & \cmark   & \cmark    & \xmark & 81.42 & 88.57 \\ \cline{2-8} 
                            &\circled{6}  & \xmark & \cmark   & \cmark    & \xmark & 75.31 & 84.05 \\ 
                            &\circled{7}  & \cmark   & \cmark   & \xmark  & \xmark & 56.87 & 67.62 \\ \Xhline{1.5pt}
  \multirow{4}{*}{Substitute} &\circled{8} & 1/L           & \cmark   & \cmark    & \cmark   & 74.68 & 84.17 \\ 
                             &\circled{9} & \cmark   & 1/L           & \cmark    & \cmark   & 1.27  & 6.07  \\ 
                            &{\small \circled{10}}  & \cmark   & \cmark   & 1/L            & \cmark   & 9.13  & 11.15 \\ 
                            &{\small \circled{11}}  & \cmark   & \cmark   & \cmark    & 1/L           & 53.17 & 59.09 \\ \Xhline{1.5pt}
  \end{tabular}
  }  
  \caption{\small Experimental results of pruning or substituting attention matrix by behavior types. $L$ is the length of input sequence such that $1/L$ is simple uniformed attention.}
  \label{tbl:pruning}
\end{minipage}~
\begin{minipage}{.52\linewidth}
  \resizebox{\linewidth}{!}{
  \begin{tabular}{c|cc|cc|cc|cc|cc}
    \Xhline{1.5pt}
    \multirow{2}{*}{id} & \multicolumn{2}{c|}{Dense}     & \multicolumn{2}{c|}{Diagonal} & \multicolumn{2}{c|}{Dense\&Vertical} & \multicolumn{2}{c|}{Vertical} & \multirow{2}{*}{Acc.} & \multirow{2}{*}{F1} \\
                                            & intra         & inter         & intra         & inter         & intra            & inter            & intra         & inter         &                       &                     \\  \Xhline{1.5pt}
                                            {\small \circled{12}}&\cmark   & \cmark   & \xmark & \cmark   & \cmark      & \cmark      & \cmark   & \cmark   & 16.96                 & 25.29               \\
                                            {\small \circled{13}}&  \cmark   & \cmark   & \cmark   & \xmark & \cmark      & \cmark      & \cmark   & \cmark   & 82.29                 & 89.91               \\
                                            {\small \circled{14}}&  \cmark   & \cmark   & \cmark   & \cmark   & \xmark    & \cmark      & \cmark   & \cmark   & 80.28                 & 87.76               \\
                                            {\small \circled{15}}&  \cmark   & \cmark   & \cmark   & \cmark   & \cmark      & \xmark    & \cmark   & \cmark   & 77.95                 & 85.86               \\  \Xhline{1.5pt}
                                            {\small \circled{16}}&\xmark & \cmark   & \cmark   & \cmark   & \xmark    & \cmark      & \xmark & \cmark   & 71.80                 & 80.22               \\
                                            {\small \circled{17}}&  \cmark   & \xmark & \cmark   & \cmark   & \cmark      & \xmark    & \cmark   & \xmark & 10.11                 & 15.57               \\
                                            {\small \circled{18}}&  \xmark & \cmark   & \xmark & \cmark   & \cmark      & \cmark      & \xmark & \cmark   & 14.83                 & 23.55               \\
                                            {\small \circled{19}}&  \cmark   & \xmark & \cmark   & \xmark & \cmark      & \cmark      & \cmark   & \xmark & 75.54                 & 84.10       \\ \Xhline{1.5pt}       
  \end{tabular}
  }
  \caption{\small Experimental results of ablating attention matrix by distance. The upper part is ablating one type. The lower part is ablating all type but remaining one type.}
  \label{tbl:pruning_distance}
\end{minipage}
\end{table}

\section{Conclusion} \label{sec:conclusion}

In this work, we propose a feasible and scalable unsupervised clustering method to classify the attention heads.
The distance based feature well captures the observations of previous works and is utilized to conduct stable unsupervised clustering.
We believe our method is helpful for attention interpretability study.
With such taxonomy, we further apply analytic experiments to explore the function of each behavior according to distance.
We are also looking forward to further analysis the behavior and function of variant patterns with probing task datasets \cite{conneau2018you} and analytic tools \cite{QiuCVPR,gan2020ptolemy} as our next plan.
Besides, there are several recent works focusing on the optimization of over-parameterized MHA mechanism \cite{michel2019sixteen,kovaleva2019revealing,guo2020accelerating}.
Our results reveal the important types and components of each type empirically.
With our explorations, we hope our insights could cast light on novel design of interpretable attention mechanism.

\vspace*{0.2cm}
\noindent\textbf{Acknowledgements*}\hspace*{0.2cm}
\small{
We thank the anonymous reviews for their thoughtful comments and suggestions.
We would like to thank Zhouhan Lin for his valuable feedback on the clustering methods and suggestions about the evaluation, and thank Jianping Zhang with whom we have inspiring discussion on the interpretability of NLP models.
This work was supported by Major Scientific Research Project of Zhejiang Lab (No. 2019DB0ZX01) and the National Natural Science Foundation of China (NSFC) grant (61702328, 61832006, and 61972247).
Jingwen Leng and Minyi Guo are corresponding authors of this paper.
}

\bibliographystyle{coling}
\bibliography{bibliography}

\appendix

\clearpage

\onecolumn

\section{All Attention Matrices} \label{sec:heatmap}

\begin{figure*}[h!]
  \centering
  \includegraphics[width=0.94\linewidth]{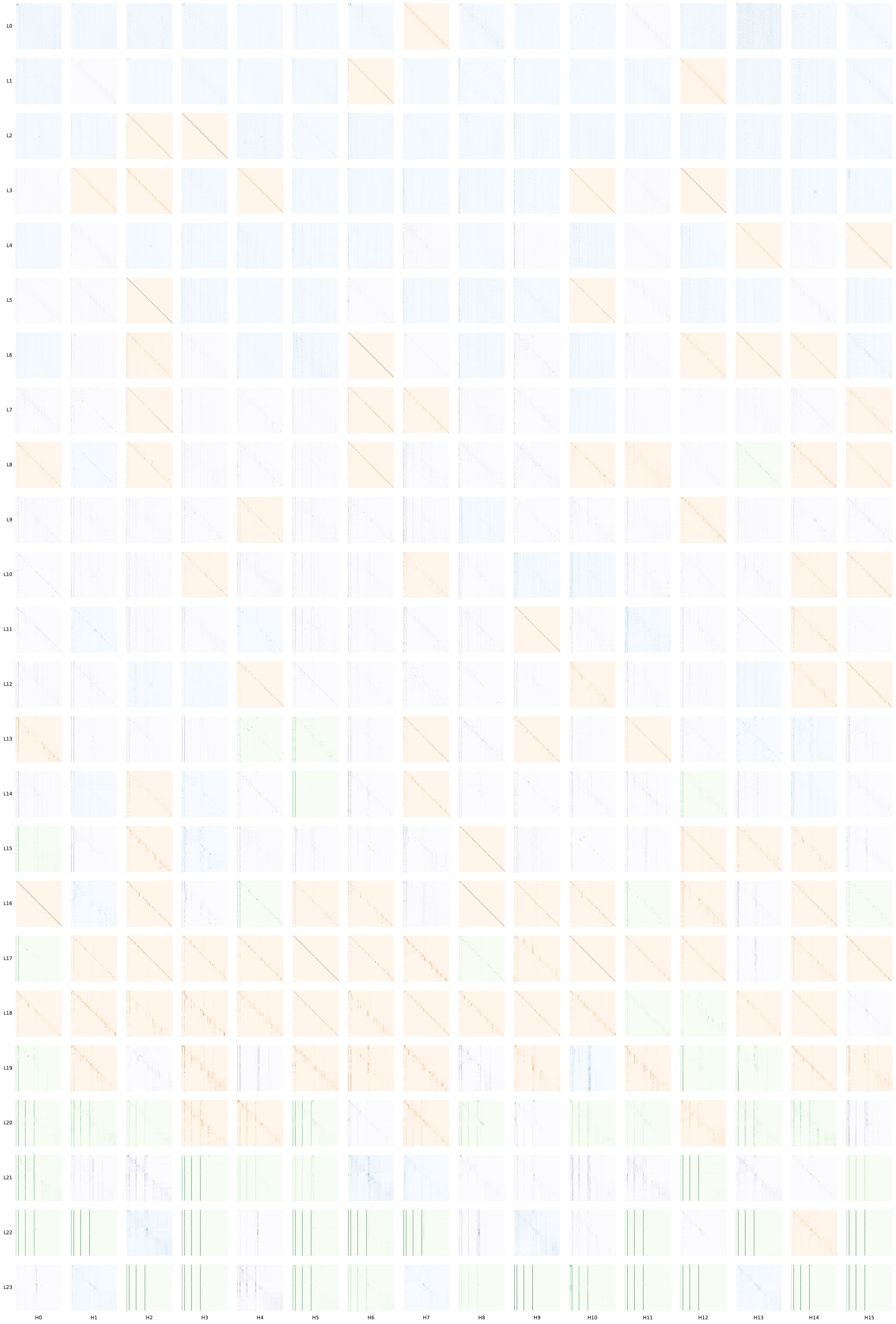}
  \caption{We visualize all attention matrices with a same randomly selected input sample as \Fig{fig:heatmap_all}.
  Attention types clustered in \Sec{sec:clustering} are distinguished by colors.}
  \label{fig:heatmap_all}
\end{figure*}

\end{document}